\documentclass[runningheads]{llncs}
\usepackage{graphicx}
\usepackage{color}
\usepackage{amsmath,amsfonts}
\usepackage{algorithmic}
\usepackage{algorithm}
\usepackage{array,multirow}
\usepackage{textcomp}
\usepackage{siunitx}
\usepackage{url}
\usepackage{booktabs}
\usepackage{verbatim}
\usepackage{subcaption}
\usepackage{hyperref}

\begin{document}
\title{A Learning-Based Model Predictive Contouring Control for Vehicle Evasive Manoeuvres\thanks{The Dutch Science Foundation NWO-TTW supports the research within the EVOLVE project (nr. 18484). European Union’s Horizon 2020 research and innovation programme under the Marie Skłodowska-Curie actions, under grant agreement (nr. 872907).}}
\titlerunning{L-MPCC for Evasive Manoeuvres}
\author{Alberto Bertipaglia\inst{1}\orcidID{0000-0003-0364-8833} \and
Mohsen Alirezaei\inst{2}\orcidID{0000-0002-2858-6408} \and
Riender Happee\inst{1}\orcidID{0000-0003-4530-8853} \and
Barys Shyrokau\inst{1} \orcidID{0000-0003-4530-8853}}
\authorrunning{A. Bertipaglia et al.}
%
\institute{Delft University of Technology, Delft 2628 CD, The Netherlands\\
\email{\{A.Bertipaglia, R.Happee, B.Shyrokau\}@tudelft.nl}\\
\and
Siemens PLM Software, Helmond 5708 JZ, The Netherlands\\
\email{mohsen.alirezaei@siemens.com}}
\maketitle              
\begin{abstract}
This paper presents a novel Learning-based Model Predictive Contouring Control (L-MPCC) algorithm for evasive manoeuvres at the limit of handling. The algorithm uses the Student-t Process (STP) to minimise model mismatches and uncertainties online. The proposed STP captures the mismatches between the prediction model and the measured lateral tyre forces and yaw rate. The mismatches correspond to the posterior means provided to the prediction model to improve its accuracy. Simultaneously, the posterior covariances are propagated to the vehicle lateral velocity and yaw rate along the prediction horizon. The STP posterior covariance directly depends on the variance of observed data, so its variance is more significant when the online measurements differ from the recorded ones in the training set and smaller in the opposite case. Thus, these covariances can be utilised in the L-MPCC's cost function to minimise the vehicle state uncertainties. In a high-fidelity simulation environment, we demonstrate that the proposed L-MPCC can successfully avoid obstacles, keeping the vehicle stable while driving a double lane change manoeuvre at a higher velocity than an MPCC without STP. Furthermore, the proposed controller yields a significantly lower peak sideslip angle, improving the vehicle's manoeuvrability compared to an L-MPCC with a Gaussian Process.
\end{abstract}
%
\section{Introduction}
\label{Introduction}
A crucial safety element of automated driving is to prove the capacity to avoid obstacles at the limit of handling. A common solution is based on Nonlinear Model Predictive Control (NMPC), which optimises the steering angle and the longitudinal force of a vehicle. However, in such scenarios, where longitudinal and lateral dynamics are coupled, the uncertainties and inaccuracies due to the tyre's non-linear behaviour pose a particularly challenging problem \cite{bertipaglia2023model,bertipaglia2024model}. Therefore, we focus on developing a Student-t Process (STP) combined with a Model Predictive Contouring Control (MPCC),  which improves the prediction model, reducing the tyre model mismatches and minimising the vehicle lateral state uncertainties. An L-MPCC based on a sparse Gaussian Process (GP) has recently been proposed for lap-time optimisation \cite{Kabzan2019Learning}. It leverages the capacity of the GP to predict the mismatches between the prediction model and the measured vehicle states: longitudinal and lateral velocity and yaw rate. Furthermore, the vehicle is constrained inside the track, tightening the track boundaries with the vehicle position uncertainty. The latter is first computed as the GP's posterior covariance, then open-loop propagated along the prediction horizon using successive linearisation similar to an Extended Kalman Filter (EKF) \cite{Kabzan2019Learning}. However, the propagated position uncertainty can increase exponentially over the prediction horizon, strongly limiting or eliminating the allowable driving area. This results in a very conservative controller. A possible alternative is to consider the uncertainty in the vehicle velocity states rather than in the vehicle position \cite{Brown2020Robust}. For instance, the NMPC cost function can be extended by the vehicle lateral velocity and yaw rate variance. Thus, the latter can be minimised to reduce the operating time in the unstable region of the vehicle. However, these variances are computed and propagated by linearising the prediction model with a constant tyre model uncertainty optimised offline. This simplification does not consider the remarkably different accuracy of the tyre model in different operating regions, and therefore, it does not reduce the prediction model mismatches in all regions.

We propose an L-MPCC based on an STP, which predicts the mismatches between the prediction model and the vehicle yaw rate and the lateral tyre forces measured by intelligent (force sensing) bearings \cite{Kerst2020Model}. The proposed controller minimises the vehicle's operating time at unstable working points, thanks to the STP posterior covariance of the tyre forces. The latter ones are used to compute and propagate the vehicle lateral state uncertainties along the prediction horizon.

The contributions of this paper are threefold. The first is the development of the L-MPCC based on an STP, which directly reduces the prediction model mismatches in the tyre model rather than in the vehicle velocity states typically used \cite{Kabzan2019Learning}. This results in successfully performing evasive manoeuvres at \SI{8.5}{\%} higher velocity than the current state-of-the-art MPCC \cite{bertipaglia2023model}. The second contribution is related to the STP, which improves the outlier resistance of the state-of-the-art GP. Furthermore, the STP posterior covariance depends on the observed measurements, providing a higher variance than a GP for operating points different than in the training set and a lower one in the opposite situation \cite{tracey2018upgrading}. The third contribution is improving the vehicle stability by reducing the sideslip angle peak of a \SI{76}{\%} during an evasive manoeuvre, thanks to the reduction of the model mismatches and the minimisation of the vehicle lateral state uncertainties. Thus, the controller decreases the time the vehicle spends in operating points close to the vehicle's handling limits.
\section{Learning-based Model Predictive Contouring Control}
The proposed L-MPCC controller is based on an MPCC for obstacle avoidance at the limit of handling \cite{bertipaglia2023model}. The prediction model is a nonlinear single-track vehicle model. The Cartesian reference system describes the vehicle kinematics, as the contouring formulation requires. The vehicle dynamics, the longitudinal ($v_x$) and lateral ($v_y$) velocity, and the vehicle yaw rate ($r$) are described as follows:
\begin{equation}
    \begin{cases}
        \dot{v}_x = \frac{\left(F_{x,\,F}\cos\left(\delta\right) - \left(F_{y,\,F} + \Delta F_{y,\,F}\right)\sin\left(\delta\right) + F_{x,\,R} - F_{drag}\right)}{m} + \left(r + \Delta r\right) v_y \\
        \dot{v}_y = \frac{\left(F_{x,\,F} \sin\left(\delta\right) + \left(F_{y,\,F} + \Delta F_{y,\,F}\right) \cos\left(\delta\right) + F_{y,\,R} + \Delta F_{y,\,R} \right)}{m} -  \left(r + \Delta r\right) v_x\\
        \dot{r} = \frac{\left(\left(F_{y,\,F} + \Delta F_{y,\,F}\right) \cos\left(\delta\right)l_f - \left(F_{y,\,R} + \Delta F_{y,\,R}\right) l_r + F_{x,\,F} \sin\left(\delta\right) l_f \right)}{I_{zz}}\\
    \end{cases}
    \label{eq:Single3}
\end{equation}
where the road-wheel angle $\left(\delta\right)$ and the longitudinal force at the front $\left(F_{x,\,F}\right)$ and rear axle $\left(F_{x,\,R}\right)$. The lateral front and rear tyre forces, respectively $F_{y,\,F}$ and $F_{y,\,R}$, are computed using validated Fiala tyre model \cite{bertipaglia2024model}. However, during an evasive manoeuvre, the prediction model inaccuracies can increase significantly, so the model mismatches of the front ($\Delta F_{y,\,F}$) and rear ($\Delta F_{y,\,R}$) lateral tyre forces and the yaw rate ($\Delta r$) are computed by an STP.

The proposed L-MPCC cost function is responsible for ensuring path tracking, maintaining the physical feasibility of the control inputs, prioritising obstacle avoidance in case of collision risk, and minimising the uncertainties of the prediction model, thus limiting the operating time in nonlinear regions. The cost function ($J$) is defined as follows:
\begin{equation}
    \begin{split}
        J =&\sum_{i=1}^{N} \Bigg( \sum_{j=1}^{N_{Obs}} \left( q_{e_{Obs, ji}} e_{Obs, ji}^2\right) + \sum_{k=1}^{N_{Edg}} \left( q_{e_{Edg, ki}} e_{Edg, ki}^2\right)+ q_{\dot{\delta}} \dot{\delta_i}^2 + q_{\dot{F}_{x}} \dot{F}_{x,\,i}^2 +\\
        & + q_{e_{Con}} e_{Con, i}^2 + q_{e_{Lag}} e_{Lag, i}^2 + q_{e_{Vel}} e_{Vel}^2 \Bigg) +\sum_{j=1}^{N_{Prob}} \left( q_{\sigma_{r}} \sigma_{r, j}^2 +q_{\sigma_{v_y}} \sigma_{v_y, j}^2 \right) 
    \end{split}
\end{equation}
where $N$ is the length of the prediction horizon, $N_{Obs}$ and $N_{Edg}$ are respectively the number of obstacles and road edges, and $q_{e_{Obs}}$, $e_{Obs}$, $q_{e_{Edg}}$ and $e_{Edg}$ are used to prioritise obstacle avoidance and keeping the vehicle away from the road edges in case of emergency \cite{bertipaglia2023model}. The tracking performance is ensured by the minimisation of the contouring ($e_{Con}$), lag ($e_{Lag}$) and velocity $e_{Vel}$ errors with their respective cost terms, i.e. $q_{e_{Con}}$, $e_{Lag}$, $q_{e_{Vel}}$. The lateral velocity and yaw rate uncertainties are minimised through the $\sigma_{v_y}$ and $\sigma_{r}$ parameters of the cost function. The $\sigma_{v_y}$ and $\sigma_{r}$ are propagated open-loop, so they can increase rapidly, potentially overpowering the other elements of the cost function. To avoid the problem, they are propagated not over the entire prediction horizon, $N=30$, but only for the reduced horizon $N_{Prob}=20$ \cite{Kabzan2019Learning}. Furthermore, the cost parameters, $q_{\sigma_{v_y}}$ and $q_{\sigma_{r}}$, are tuned not to exceed the cost related to obstacle avoidance prioritisation. Therefore, the proposed controller pushes the vehicle to work at operating points close to normal driving. However, it allows the vehicle to drive at operating points close to the handling limits if it helps avoid a collision.

\section{Student-T Process \& Uncertainty Propagation}
\label{Student}
This study uses an STP as a stochastic process with a Laplace inference to estimate model mismatches and uncertainties for two reasons. First, the Student-t distribution allows the algorithm to define the level of Kurtosis, reducing the influence of the outliers and improving the accuracy of the predictions \cite{shah2014student,vanhatalo2009gaussian}. The robustness to outliers is an essential property for L-MPCC, which relies on online measurements with sensor noise. For instance, a single outlier can highly influence the GP prediction; pushing the posterior means far away from the level of the other observations, which is not the case in an STP \cite{vanhatalo2009gaussian}. The second reason is that the STP posterior covariance directly depends on the observed measurements and not only on the location of the observed measurements. This implies that the covariance increases when the measurements vary more than expected, i.e., a higher difference between training and test sets, and vice-versa decreases when the difference is lower \cite{tracey2018upgrading,shah2014student}. Once again, this is an essential property for an L-MPCC, which can work in conditions with a high discrepancy from the training scenario. Regarding the kernel selection, this work opts for the automatic relevance determination Matérn $5/2$ function due to its generalisation capabilities and interpretability properties. The STP is implemented using the GPML Matlab Code version 4.2 \cite{rasmussen2010gaussian}.

We predict the  $\Delta F_{y,\,F}$, $\Delta F_{y,\,R}$ and $\Delta r$ model mismatches, rather than the errors in $v_x$ and $v_y$ as usually done in literature \cite{Kabzan2019Learning} because it allows the reduction of the mismatches in a proactive way. It also directly targets the source of the errors, i.e., the tyre forces, and not the states that depend on them. This is possible thanks to the availability of tyre lateral forces at the front and rear axle, measured by the intelligent bearings \cite{Kerst2020Model}, the vehicle yaw rate and the control inputs $F_x$ and $\delta$. Thus, we assume that the discrepancies are independent of the vehicle position \cite{Kabzan2019Learning} and that the velocity states, not measurable through the standard sensor setup for passenger vehicles, are entirely dependent on the tyre forces. The training data is derived from the algebraic difference between the measurements and the nominal model predictions. The training set is composed of 2 double lane changes at respectively \SI{55}{km/h} and \SI{80}{km/h}, and a double lane change with collision avoidance prioritisation at \SI{55}{km/h}. The manoeuvres are selected to identify the mismatches in the linear and nonlinear regions of the vehicle. The test set comprises a double lane change at \SI{60}{km/h} and the same manoeuvre with collision prioritisation. Despite the high prediction model accuracy and the reduction of the model mismatches, some unmodeled effects can still influence the prediction model. These are captured by the STP posterior covariance for the front and rear lateral tyre force, respectively $\sigma_{F_{y,\, F}}$ and $\sigma_{F_{y,\, R}}$. Thus, the tyre forces are not only evaluated as a single point but also as a distribution \cite{Brown2020Robust}. The STP variance and degrees of freedom are transformed and simplified into a Gaussian variance to be able to propagate it through the prediction horizon and relate them to the lateral states related to vehicle stability, i.e. $v_y$ and $r$, using a procedure similar to an EKF \cite{Kabzan2019Learning,Brown2020Robust}.
\section{Results}
\label{Results}
\begin{figure*}[!t]
    \centering
    \begin{subfigure}{1\columnwidth}
        \centering
        \includegraphics[height=2.8cm, keepaspectratio]{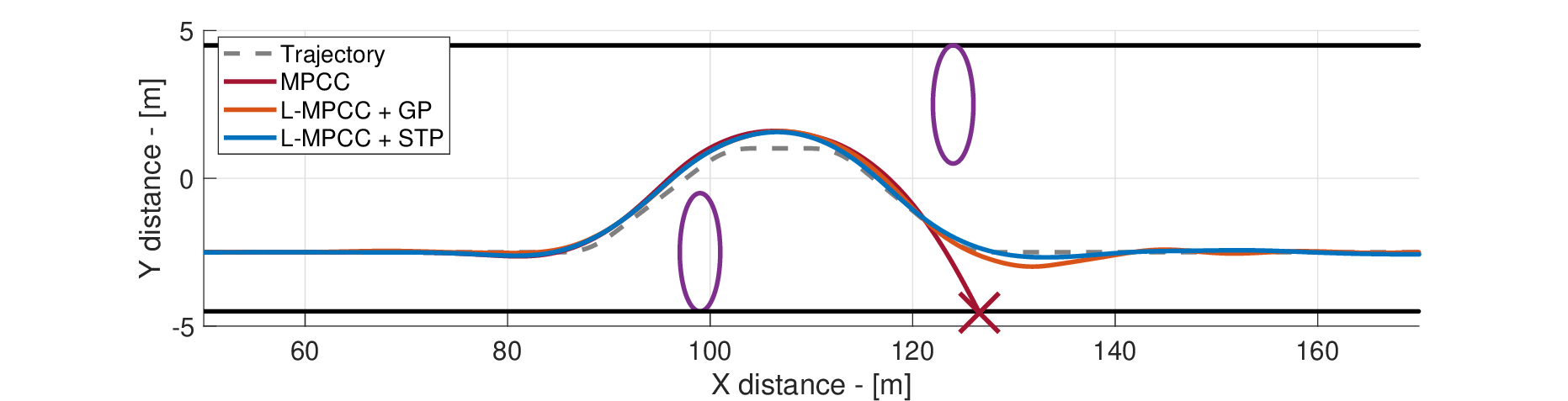}
        \caption{}
        \label{fig:Trajectory}
    \end{subfigure}\\
    \begin{subfigure}{0.32\columnwidth}
        \centering
        \includegraphics[width=1\columnwidth]{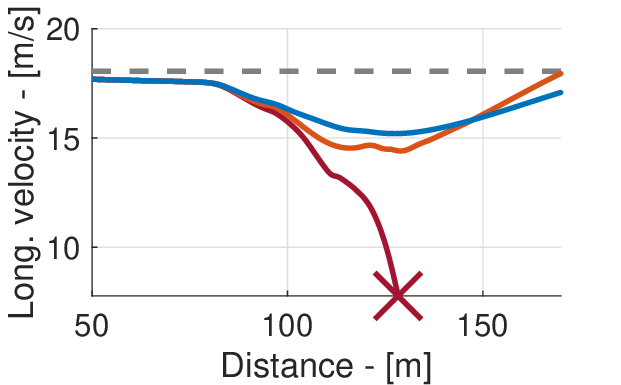}
        \caption{}
        \label{fig:vx}
    \end{subfigure}
    \begin{subfigure}{0.32\columnwidth}
        \centering
        \includegraphics[width=1\columnwidth]{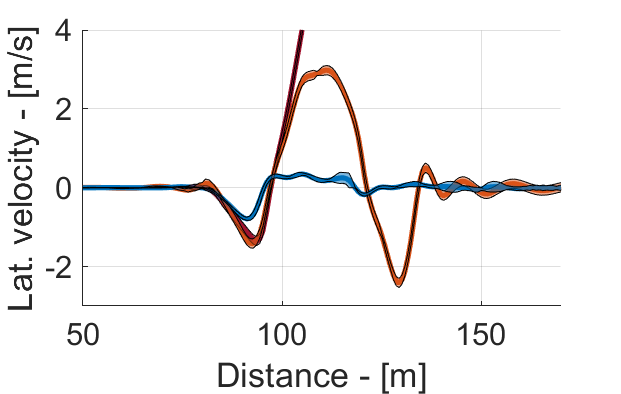}
        \caption{}
        \label{fig:vy}
    \end{subfigure}
    \begin{subfigure}{0.32\columnwidth}
        \centering
        \includegraphics[width=0.75\columnwidth]{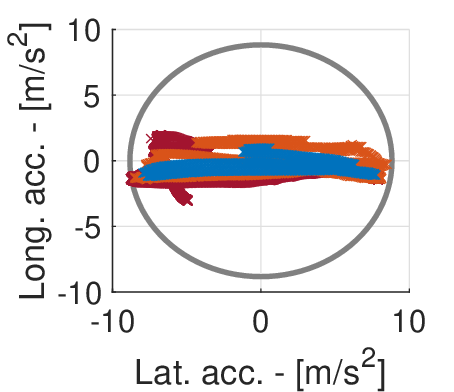}
        \caption{}
        \label{fig:GG}
    \end{subfigure}\hfill\\
    \begin{subfigure}{0.32\columnwidth}
        \centering
        \includegraphics[width=1\columnwidth]{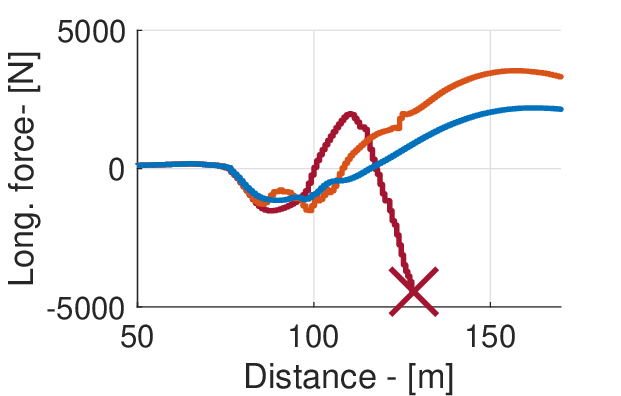}
        \caption{}
        \label{fig:Fx}
    \end{subfigure}
    \begin{subfigure}{0.32\columnwidth}
        \centering
        \includegraphics[width=1\columnwidth]{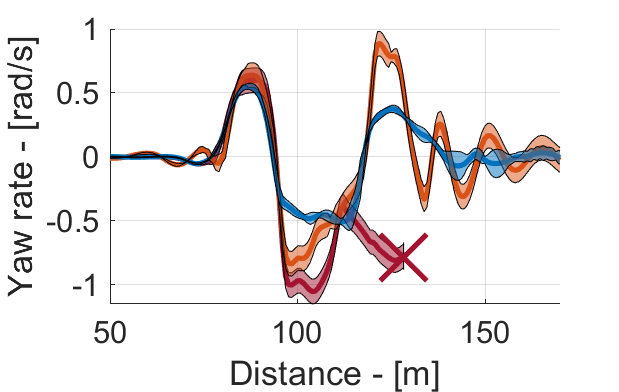}
        \caption{}
        \label{fig:yawrate}
    \end{subfigure}
    \begin{subfigure}{0.32\columnwidth}
        \centering
        \includegraphics[width=1\columnwidth]{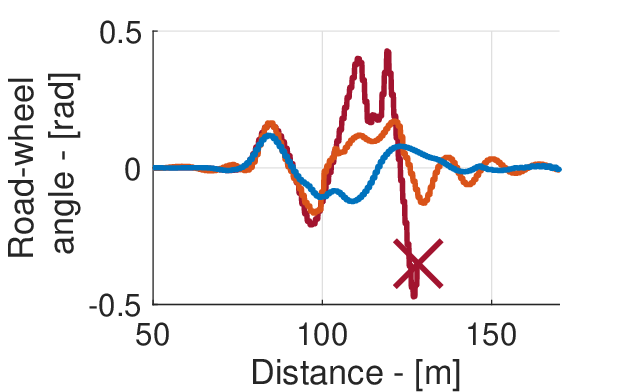}
        \caption{}
        \label{fig:delta}
    \end{subfigure}\hfill
    \caption{States and control inputs in a double lane change.}
    \label{fig:DoubleLane}
\end{figure*}
Fig. \ref{fig:DoubleLane} shows the performance of the evaluated controllers in a double lane change with collision avoidance prioritisation at \SI{65}{km/h}. Fig. \ref{fig:Trajectory} shows that only the learning-based controllers can successfully avoid the two obstacles and stay inside the road track. Vice versa, the MPCC baseline \cite{bertipaglia2023model} leaves the track on the right side at \SI{130}{m}. The improvement is due to reduced model mismatches in the L-MPCC. For instance, the proposed L-MPCC+STP reduces the root mean square of the $F_{y,\, F}$, $F_{y,\, R}$ and $r$ mismatches by respectively \SI{33.14}{\%}, \SI{24.85}{\%} and \SI{60.61}{\%} compared to the MPCC and further \SI{20.14}{\%}, \SI{11.85}{\%} and \SI{61.33}{\%} with respect to the L-MPCC+GP. This proves the importance of implementing an STP rather than a GP to predict the model mismatches. The enhanced performance of the proposed controller is also visible from the reduction of the vehicle lateral velocity peak; see Fig. \ref{fig:vy}. Furthermore, the L-MPCC+STP can reduce the lateral velocity and, consequently, the sideslip angle peak by \SI{76}{\%}, maintaining an overall higher velocity during the manoeuvre than the L-MPCC+GP. Figs. \ref{fig:vy} and \ref{fig:yawrate} show that the vehicle sideslip angle reduction and the increased operating time in the stable vehicle behaviour are not only due to the lower model mismatches but also to the reduced uncertainties for the vehicle lateral states. However, it has a lower vehicle longitudinal acceleration when the obstacle avoidance manoeuvre is over to achieve the desired velocity, Fig. \ref{fig:Fx}.
\section{Conclusions}
\label{Conclusions}
This paper presented a novel Learning-based Model Predictive Contouring Control based on a Student-t Process for evasive manoeuvres with an online minimisation of model mismatches and uncertainties. In a high-fidelity simulation environment, we demonstrate that our proposed controller successfully avoids obstacles, keeping the vehicle stable while driving a double-lane change manoeuvre at a higher velocity compared to a non-learning-based baseline. Furthermore, the proposed controller reduces the peak of vehicle sideslip angle by \SI{76}{\%}. The performance enhancement is due to the properties of the Student-t Process, which can further reduce the prediction model mismatches and better capture the vehicle state uncertainties. Future work involves the proposed controller implementation on rapid prototyping hardware to prove its real-time capability.
\bibliographystyle{splncs04}
\bibliography{references.bib}
\end{document}